\documentclass[conference]{IEEEtran}
\IEEEoverridecommandlockouts
\usepackage{cite}
\usepackage{amsmath,amssymb,amsfonts}
\usepackage{algorithmic}
\usepackage{graphicx}
\usepackage{textcomp}
\usepackage{xcolor}
\usepackage{multirow}
\usepackage{listings}
\usepackage{caption}
\usepackage[numbers,sort&compress,sectionbib]{natbib} 
\usepackage{booktabs}
\usepackage{colortbl}
\definecolor{lightgreen}{RGB}{210, 255, 210}
\usepackage{float}  
\usepackage{makecell}

\newcommand\sr[1]{\textbf{\textcolor{purple}{SR: #1}}}
\def\BibTeX{{\rm B\kern-.05em{\sc i\kern-.025em b}\kern-.08em
    T\kern-.1667em\lower.7ex\hbox{E}\kern-.125emX}}
\usepackage{url}       
\usepackage{footnote}  
\usepackage{hyperref}

\begin{document}

\title{SinLlama - A Large Language Model for Sinhala}



\author{
\IEEEauthorblockN{%
H.W.K.Aravinda\IEEEauthorrefmark{1},
Rashad Sirajudeen\IEEEauthorrefmark{1},
Samith Karunathilake\IEEEauthorrefmark{1},
\\
Nisansa de Silva\IEEEauthorrefmark{1},
Rishemjit Kaur\IEEEauthorrefmark{2}\IEEEauthorrefmark{3},
Arshdeep Singh Bhankhar\IEEEauthorrefmark{3},
Surangika Ranathunga\IEEEauthorrefmark{4}
}

\IEEEauthorblockA{\IEEEauthorrefmark{1}Dept.\ of Computer Science \& Engineering, University of Moratuwa, Sri Lanka.\\
\texttt{\{aravinda.20, rashad.20, samith.20, sandarekaw, NisansaDds\}@cse.mrt.ac.lk}}

\IEEEauthorblockA{\IEEEauthorrefmark{2}Central Scientific Instruments Organisation, Academy of Scientific and Innovative Research, Chandigarh, India.\\
\texttt{rishemjit.kaur@csio.res.in}}

\IEEEauthorblockA{\IEEEauthorrefmark{3}IIT Ropar Technology \& Innovation Foundation (iHub - AWaDH), India \& CSIR-CSIO Chandigarh, India\\
\texttt{abhankhar\_be20@thapar.edu}}

\IEEEauthorblockA{\IEEEauthorrefmark{4}School of Mathematical and Computational Sciences, Massey University, Auckland, New Zealand.\\
\texttt{s.ranathunga@massey.ac.nz}}
}

\maketitle






\begin{abstract}
Low-resource languages such as Sinhala are often overlooked by open-source Large Language Models (LLMs). Therefore, it is imperative that the existing LLMs are further trained to cover such languages. In this research, we extend an existing multilingual LLM (Llama-3-8B) to get a better coverage for Sinhala. We enhanced the LLM tokenizer with Sinhala specific vocabulary and performed continual pre-training on a 10 million sentence Sinhala corpus, resulting in the \textbf{SinLlama} model. This is the very first decoder-based open-source LLM with explicit Sinhala support. When SinLlama was instruction fine-tuned for three text classification tasks, it outperformed base and instruct variants of Llama-3-8B by a significant margin.
\end{abstract}


\section{Introduction}
Large Language Models (LLMs) such as ChatGPT have significantly advanced the field of Natural Language Processing (NLP) for languages with abundant training data and linguistic resources. However, their support for low-resource languages (LRLs) is significantly low. While some recent successful open-source LLMs such as Llama~\cite{grattafiori2024Llama} and Gemma~\cite{team2024gemma} claim to have multilingual support, such multilingual LLMs (multiLLMs) continue to under-perform on LRLs~\cite{zhang-etal-2023-dont}. Given that these LLMs are being created by large tech corporations, it is fair to assume that the decision on which languages to include in an LLM is driven by the perceived economic benefits in a global scale. As such, there is a risk of LRLs continue to be neglected, thus further increasing the digital divide between communities. 

Sinhala is an Indo-Aryan language, which is only being used by a population of about 20 million in the island nation of Sri Lanka~\cite{de2019survey}. According to~\citet{ranathunga2022some}'s language categorization that uses the category definition of~\citet{joshi-etal-2020-state}, Sinhala is a low-resource language. There has been previous research that resulted in word embeddings and encoder-based language models for Sinhala~\cite{dhananjaya2022bertifying}. However, techniques based on such models have become out-dated due to the demonstrated potential of LLMs. However, Sinhala, similar to many other LRLs, has not been considered during pre-training of any of the popular open-source LLMs such as Llama and Gemma released by the tech giants. While Sinhala is included in  MaLA-500~\cite{lin2024mala} and Aya-101~\cite{ustun2024aya} released by research teams, these models have not been able to outperform models such as Llama. On the other hand, MaLA-500 and Aya-101 models are large in size, incurring in large computer memory requirements when used for downstream tasks. 

To address these limitations, we present a focused adaptation of a state-of-the-art multiLLM to better support Sinhala. Our approach involves two key stages: (1) enhancing the model's tokenizer by merging Sinhala-specific tokens into its vocabulary to improve subword segmentation; and (2) conducting continual pretraining using a 10 million sentence corpus that includes cleaned and diverse Sinhala text, ensuring deeper linguistic representation.

We first carried out an investigation on the available open-source LLMs to select the best performing LLM for Sinhala. Llama-3 emerged as the most promising LLM. Then we further pre-trained the Llama-3-8B model as mentioned above.  We term the resulting model, \textbf{SinLlama}\footnote{\url{https://huggingface.co/polyglots/SinLlama_v01}}, which is publicly released. In order to evaluate the performance of SinLlama against the original Llama-3, we fine-tuned both with data from three Sinhala benchmark datasetss~\cite{wang2018glue}: writing-style classification, news categorization and sentiment analysis. The results show that when fine-tuned on Sinhala task-specific data, SinLlama significantly outperforms both fine-tuned Llama-base and Llama-instruct versions.  

\section{Related Work}
\label{sec:relatedwork}
\subsection{Large Language Models (LLMs)}
Pre-trained Language Models (PLMs), created using the Transformer architecture~\cite{vaswani2017attention}, have significantly advanced the NLP landscape. Among them, decoder-only models such as ChatGPT have become especially popular due to their strong performance across a multitude of tasks. Open-source large-scale decoder models, known as LLMs (e.g., LLaMA, Mistral~\cite{jiang2023mistral7b}, Gemma), are also trained on larger datasets and offer performance comparable to commercial systems. Training an LLM has two primary stages: training a tokenizer and self-supervised training with an unlabeled raw text corpus. The latter step is termed \textit{pre-training}, and the resulting model is termed a \textit{base} model. However, such base models lack task-specific capabilities. Therefore, these base models should be fine-tuned with task-specific instruction data, which is termed \textit{fine-tuning} or \textit{instruct-tuning}.  The aforementioned open-source LLMs are available in both base and instruct versions.

\subsection{Empirical Studies on LLM Performance for LRLs}
The evaluation of LLMs across multiple languages reveals significant performance disparities between high-resource and LRLs.~\citet{ahuja2023mega}'s MEGA framework conducted the first comprehensive benchmarking of LLMs on 16 NLP tasks across 70 diverse languages (including Sinhala) for comparing the performance of different models on XLSUM dataset. They showed that while models such as GPT-4 performed well in high-resource languages, they struggle with LRLs, especially those with unique morphological structures and/or scripts.

Recent surveys of LLMs underscore the influence of pretraining data, language typology, and evaluation benchmarks on model performance across languages. \citet{chang2024survey} conducted a comprehensive review of over 200 LLMs, highlighting that multilingual performance is significantly affected by the diversity and representativeness of the pretraining corpus, with underrepresented languages, especially those with complex morphology or distinct scripts, often exhibiting degraded results. The survey also emphasizes the limitations of current multilingual benchmarks, which are either narrow in task scope or limited in linguistic coverage, leading to an incomplete understanding of LLM capabilities in low-resource languages.

\subsection{Extending LLMs for New Languages}
\label{subsec:extending}
There are several strategies to extend an LLM to an unseen language. 
\begin{itemize}
    \item \textbf{Continual Pre-training:} This involves further training an existing LLM on text data from the target language to improve its language understanding and generation capabilities. Continued pre-training can improve alignment with the new language, but may risk degrading performance in previously learned languages if not carefully managed~\cite{cossu2024continual}.
    
    \item \textbf{Vocabulary Extension:} Adding new tokens specific to the target language to the tokenizer's vocabulary improves encoding efficiency and model performance. Simple vocabulary extension combined with continued pre-training has been shown to be effective in various languages~\cite{toraman2024llamaturk}.
    
    \item \textbf{Fine-tuning:} Instruction tuning or task-specific fine-tuning on data in the new language helps the model adapt to language-specific nuances and tasks. Techniques such as translation-assisted chain-of-thought fine-tuning (TaCo)~\cite{upadhayay2023taco} have been proposed to improve cross-lingual transfer for LRLs.
    
    \item \textbf{Preference Optimization:} This method optimizes the model based on human preference data, which can be scarce for LRLs. Preference optimization helps align model output more closely with human expectations in the new language context.

    \item \textbf{In-context Learning:} Providing relevant examples to LLMs during inference is known as in-context learning or few-shot learning. In-context learning has shown to work in the context of LRLs as well~\cite{moslem2023adaptive}.
    
    \item \textbf{Leveraging  Lexicons:} Structured language-specific linguistic resources such as lexicons can be integrated with LLMs to boost their performance for LRLs~\cite{koto2024zero}. Unlike zero-shot methods, which often fail in domain-specific settings, few-shot prompts enhance both accuracy and efficiency, marking a shift toward more adaptive strategies.
\end{itemize}


\subsection{Sinhala Language Computing}
While there has been research on implementing Sinhala language computing tools, most of it has been done using rule-based, statistical or early-generation deep learning techniques~\cite{de2019survey}. In the NLP field, these techniques have long become obsolete. Nevertheless, there have been some attempts to keep Sinhala NLP up-to-date.~\citet{lakmal2020word} presented Word2Vec and FastText embeddings for Sinhala and carried out a detailed investigation.~\citet{dhananjaya2022bertifying} built the encoder-based language model SinBERT, by fine-tuning an encoder-based Transformer~\cite{liu2019roberta} with 15.7 million Sinhala sentences. 

\section{Model Selection}\label{AA}

In the initial phase of our research, we aimed to identify the most suitable open-source multiLLM, focusing on LRLs. The first round of was carried out referring to empirical research that compared open-source LLMs.  This resulted in the following models: Llama-3-8B~\cite{grattafiori2024Llama}, MaLA-500~\cite{lin2024mala}, Aya-101~\cite{ustun2024aya}, PolyLM~\cite{wei2023polylm}, XGLM,~\cite{li2024eliciting} Falcon~\cite{almazrouei2023falcon}, BLOOMZ~\cite{muennighoff2022crosslingual}, Phi3~\cite{abdin2024phi}, Mistral~\cite{jiang2023mistral7b}, OPT~\cite{zhang2022opt}, Aya-23~\cite{ustun2024aya}, and BigTrans~\cite{yang2305bigtranslate}. These models are listed in Table~\ref{tab:model-details}. Based on criteria including multilingual capability, parameter size (fewer than 8 billion parameters)\footnote{Model size has a direct relationship with the memory requirement in using an LLM.}, Sinhala language representation, and performance metrics, we selected Llama-3-8B \cite{grattafiori2024Llama} 
for further experimentation.

\begin{table}[h]
\caption{Model Details and Sinhala Language Support\iffalse\sr{mention the number of languagesin each model}\fi}
\centering
\label{tab:model-details}

\begin{tabular}{|l|r|c|r|}
\hline
Model & Parameters & \makecell{Official\\ Sinhala Support} & \makecell{Number of\\languages}\\ \hline

Llama-3-8B & 8B & No & 8\\ \hline
MaLA-500 & 10.7B & Yes &  534 \\ \hline
Aya-101 & 13B & Yes & 101\\ \hline
PolyLM-13B & 13B & No & 18\\ \hline
XGLM-7.5B & 7.5B & Yes & 30 \\ \hline
Falcon-7B & 7B & No & 15\\ \hline
BLOOMZ-7.1B & 7.1B & No & 46 \\ \hline
Phi-3-mini & 3.8B & No & 14 \\ \hline
Mistral-7B & 7B & No & 10 \\ \hline
OPT-66B & 66B & No & 5 \\ \hline
Aya-23-8B & 8B & No & 23\\ \hline
BigTrans-13B & 13B & Yes & 102  \\ \hline

\end{tabular}
\vspace{-0.5em}
\end{table}

{Llama-3-8B}~\cite{grattafiori2024Llama} officially supports 8 languages and is pre-trained on a large dataset over 15 trillion data tokens, sourced from publicly available Internet materials. It has 128K context length. Lalama-3-8B outperformed its counterparts such as Mistral 7B and Gemma 7B on common benchmarks such as MMLU~\cite{hendrycks2020measuring}, ARC~\cite{clark2018think}, DROP~\cite{dua2019drop} and GPQA~\cite{rein2024gpqa}.

\section{Dataset}

\subsection{Pre-training Data}
Pre-training of an LLM has to be done with a substantially large dataset, in order to guarantee it learns the language-specific information.  Therefore, we created a dataset by combining MADLAD-400~\cite{kudugunta2023madlad} and CulturaX~\cite{nguyen2023culturax} for the Sinhala language\footnote{The dataset created by~\citet{dhananjaya2022bertifying} is no longer accessible. According to~\citet{ranathunga2024shoulders}, this is a common problem in LRL domain.}. MADLAD-400 is a manually audited, general-domain 3 Trillion token monolingual dataset based on CommonCrawl~\cite{patel2020introduction}, spanning 419 languages. CulturaX is a  multilingual dataset with 6.3 Trillion tokens in 167 languages. Although both have been cleaned, there were some entries that contained non-Sinhala words. Because of that, heuristics-based filtering~\cite{fernando2025improving} was applied to remove such noisy sentences. These include rules such as minimum sentence length, non-Sinhala sentence removal, exclusion of sentences with irregular punctuation, and elimination of URLs and numeric prefixes.
For the pre-training, we combined both datasets and removed duplicates. Our final dataset contains 10,730,154 (10.7M) Sinhala sentences, comprising 303,958,959 (303.96M) tokens. This dataset is publicly released\footnote{\url{https://huggingface.co/datasets/polyglots/MADLAD_CulturaX_cleaned}}. 

\subsection{Fine-tuning Data}
\label{sec:fine-tune-data}
For the supervised fine-tuning experiments, we focused on Sinhala text classification tasks. We selected three publicly available datasets: News Category Classification~\cite{de2015sinhala}, Sentiment Analysis~\cite{ranathunga2021sentiment}, and Writing Style Classification~\cite{dhananjaya2022bertifying}.These datasets collectively represent a diverse range of linguistic features and styles, making them suitable for evaluating the effectiveness of LLMs for Sinhala.

Since the original datasets were not in a directly usable format for fine-tuning, several preprocessing steps were performed. The Sentiment classification dataset, originally in XML format, required extraction of relevant fields: content within the \verb|<phrase>| tag was used as the input text, and the \verb|<sentiment>| tag provided the corresponding label. For other datasets, we only applied standard text cleaning procedures such as deduplication, removal of malformed entries, and whitespace normalization.

To enable robust evaluation, we manually split each dataset into training, validation, and test sets using an 80:10:10 ratio. We employed stratified sampling to ensure that each split preserved the original label distribution, thereby preventing skewed performance metrics. The final dataset sizes used for fine-tuning are shown in Table~\ref{tab:dataset-size}.

\begin{table}[h]
\caption{Dataset sizes for Text Classificiation Task}
\centering
\label{tab:dataset-size}

\begin{tabular}{|l|r|r|r|r|r|}
\hline
                          Dataset   & Train   & Test & Validation  \\ \hline

Writting style classification~\cite{dhananjaya2022bertifying}      &  10010 &  1252 & 1252 \\ \hline 
Sentiment analysis~\cite{ranathunga2021sentiment}   & 7238 	&  905 & 905 \\ \hline
Sinhala news category classification~\cite{de2015sinhala}  & 2661 &	333 &	333 \\ \hline

\end{tabular}
\vspace{-0.5em}
\end{table}

\section{Continual Pretraining to Build SinLlama}
We used vocabulary extension and continuous pre-training (see Section~\ref{subsec:extending}) on Llama-3-8B to create SinLlama. We trained a tokenizer using the pre-training corpus using tiktoken\footnote{\url{https://github.com/openai/tiktoken}}, and merged this tokenizer with the Llama 3 tokenizer. We used this extended tokenizer and performed continual pretraining using the codebase from Chinese-Llama~\cite{cui2023efficient}. The only modification made to the original Chinese-Llama hyperparameters was the reduction of the \texttt{block\_size} from 1024 to 512\footnote{Larger block sizes cause memory allocation issues on the GPUs we used, preventing successful model loading.}. 

\section{Task-Specific Finetuning}
\subsection{Prompt Selection}

\begin{figure}[H]
    \centering
    \includegraphics[width=\linewidth]{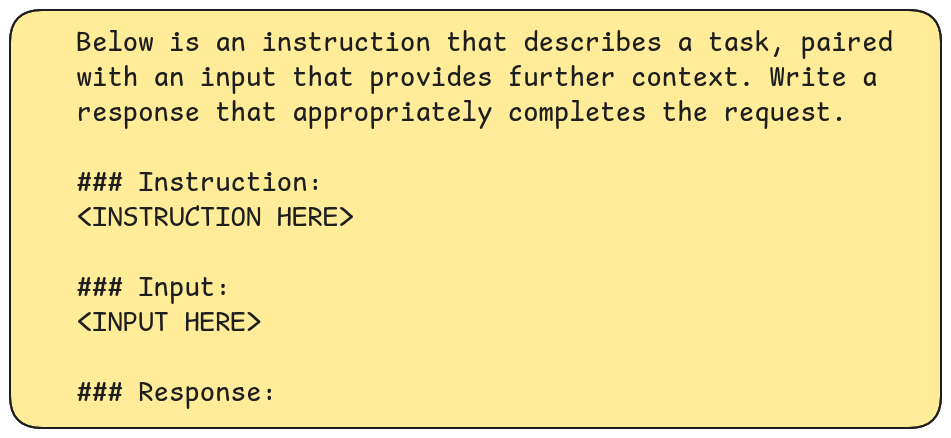}
    \caption{Prompt template for supervised fine-tuning}
    \label{fig:prompt}
\end{figure}

For both inference and supervised fine-tuning, it was essential to select the most effective prompt templates from the available datasets. We adopted the prompt format introduced by the Alpaca Instruct Template~\cite{chen2023monolingual} as in Fig.~\ref{fig:prompt}. To identify the most suitable prompts for each classification task, we referred to the work by~\citet{ahuja2023mega}, which provided well-designed prompts for several tasks.







\subsection{Fine-tuning}

\begin{table*}[!htb]
\caption{Performance Comparison Across Tasks}
\centering
\label{tab:combined-results}
\begin{tabular}{|l|ccc|ccc|ccc|}
\hline
\textbf{Model} & \multicolumn{3}{c|}{\textbf{Writing Style}} & \multicolumn{3}{c|}{\textbf{News Category}} & \multicolumn{3}{c|}{\textbf{Sentiment Tagger}} \\
\cline{2-10}
 & \textbf{Precision} & \textbf{Recall} & \textbf{F1} & \textbf{Precision} & \textbf{Recall} & \textbf{F1} & \textbf{Precision} & \textbf{Recall} & \textbf{F1} \\
\hline
Llama-3-8B base& 33.091 & 20.927 & 24.504 & 23.904 & 18.919 & 19.031 & 41.942 & 38.011 & 36.285 \\
Llama-3-8B base - Finetuned & 71.060 & 38.179 & 49.451 & 64.770 & 63.664 & 61.136 & 61.110 & 61.215 & 59.353 \\
Llama-3-8B instruct & 50.782 & 50.719 & 48.755 & 31.866 & 30.330 & 29.219 & 48.451 & 36.796 & 31.257 \\
Llama-3-8B instruct - Finetuned & 70.996 & 35.942 & 42.256 & 50.883 & 50.450 & 47.812 & 68.856 & 68.729 & 68.784 \\
SinLlama & 33.315 & 4.393 & 7.531 & 44.166 & 19.520 & 15.614 & 47.153 & 25.856 & 22.689 \\
\rowcolor{lightgreen}
\textbf{SinLlama - Finetuned} & \textbf{85.906} & \textbf{52.157} & \textbf{58.893} & \textbf{89.033} & \textbf{86.787} & \textbf{86.402} & \textbf{75.246} & \textbf{70.055} & \textbf{72.471} \\
\hline
\end{tabular}
\vspace{-0.5em}
\end{table*}

We fine-tuned SinLlama and the Llama-3-8B base model separately with the training split of each dataset mentioned in Section~\ref{sec:fine-tune-data}, creating a set of task-specific fine-tuned models. 
We used LoRA fine-tuning~\cite{hu2022lora}, which enables faster training and lower memory usage. 

\begin{figure}[H]
    \centering
    \includegraphics[width=0.46\textwidth]{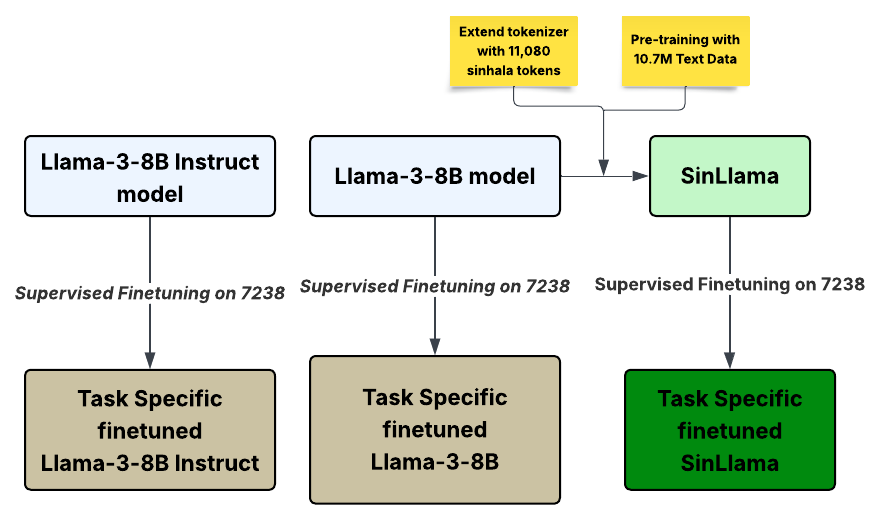} 
    \caption{An example experimental setup for the Writing Style Classification dataset.\iffalse\sr{correct spellin is Llama. fix all}\fi}
    \label{fig:experiment_setup}
\end{figure}

Figure~\ref{fig:experiment_setup} shows an example experimental setup for one dataset; we followed a similar process for other datasets.

\section{Evaluation}
We used Llama-3-8B base original model, Llama-3-8B-instruct original model, and their fine-tuned (with our data) versions, as the baselines. 
From the results of table \ref{tab:combined-results}, we can identify that the fine-tuned SinLlama model performs best at each of the task with a significantly high margin. When the Llama-3-8B base model is fine-tuned with our dataset, there is a significant gain over the base model. However, this result is far below the fine-tuned SinLlama model. Fine-tuned Llama-3-8B instruct model also outperforms its un-fine-tuned version. However, it falls below the fine-tuned Llama-3-8B base model, which is surprising. Overall, the writing style classification task seems to be the most difficult for the LLMs. Because the input field of that dataset is much larger than those of the other two datasets.

\section{Conclusion}
Although Sinhala has been included in several multilingual pre-trained large language models, it was not intentionally targeted during the pre-training phase. In this study, we introduce SinLlama, the first-ever large language model pre-trained specifically for Sinhala. 
Our results demonstrate that fine-tuning SinLlama with Sinhala task-specific data results in significant gains compared to fine-tuning Llama-3-8B base and instruct models. In the 
future, we plan to conduct experiments on several fine-tuning strategies and preference optimization methods.

\section*{Acknowledgment}
We thank {CSIR - Central Scientific Instruments Organization}, India, and {Emojot (Pvt) Ltd}  for providing the computing resources.

\bibliographystyle{IEEEtranN}
\bibliography{merconbib}           

\begin{appendices}
\label{Appendix}
\section{Prompts Used}
All of the prompts used for the benchmark process referred to the prompts given from the \citet{ahuja2023mega}. Selected prompts are given below.

\subsection{Writing-style-classification}
Feature columns in the dataset are ``\textbf{comments}", ``\textbf{labels}".

\begin{lstlisting}[basicstyle=\ttfamily\footnotesize]
You are an NLP assistant whose purpose is to classify 
Sinhala comments into predefined categories. The 
categories are as follows: ACADEMIC, CREATIVE, NEWS,
BLOG. Given a Sinhala comment, your task is to 
determine which category it belongs to. Choose one
category: ACADEMIC, CREATIVE, NEWS, or BLOG. Answer 
must match exactly in capitalization and formatting. 
Comment: {comment} 
Answer:
\end{lstlisting}

\subsection{Sentiment-tagger}
The original dataset consist of XML tags. After the processing we created a dataset with the following features.
Feature columns in the dataset are ``\textbf{context}", ``\textbf{text}", ``\textbf{sentiment}".

\begin{lstlisting}[basicstyle=\ttfamily\footnotesize]
You are an NLP assistant whose purpose is to solve 
Sentiment Analysis problems. Sentiment Analysis is 
the task of determining whether the sentiment, 
opinion or emotion expressed in a textual data is. 
Does the following Sinhala sentence have a POSITIVE,
NEGATIVE or NEUTRAL sentiment? {text}. Give only 
the POSITIVE, NEGATIVE, or NEUTRAL as the answer in 
one word.
\end{lstlisting}

\subsection{Sinhala-News-Category-classification}
Feature columns in the dataset are ``\textbf{comments}", ``\textbf{labels}".

\begin{lstlisting}[basicstyle=\ttfamily\footnotesize]
You are an NLP assistant whose purpose is to classify
Sinhala comments into predefined categories. The 
categories are as follows: (Political: 0, Business: 1, 
Technology: 2, Sports: 3, Entertainment: 4) Given a 
Sinhala comment, your task is to determine which 
category it belongs to. Please only give the label 
as a number (0, 1, 2, 3, or 4) that best represents 
the comment's category. comment: {text} answer:
\end{lstlisting}

\end{appendices}

\end{document}